# Beyond Least Squares: Robust Regression Transformer (R2T)

Roman Gutierrez
Vistendo Inc
romancg@vistendo.com

Tony Kai Tang
Vistendo Inc.
ttang@vistendo.com

Isabel Gutierrez
Vistendo Inc.
isabelg@vistendo.com

## Abstract

Robust regression techniques rely on least-squares optimization, which works well for Gaussian noise but fails in the presence of asymmetric structured noise. We propose a hybrid neural-symbolic architecture where a transformer encoder processes numerical sequences, a compression NN predicts symbolic parameters, and a fixed symbolic equation reconstructs the original sequence. Using synthetic data, the training objective is to recover the original sequence after adding asymmetric structured noise, effectively learning a symbolic fit guided by neural parameter estimation. Our model achieves a median regression MSE of 6e-6 to 3.5e-5 on synthetic wearable data, which is a 10-300 times improvement when compared with ordinary least squares fit and robust regression techniques such as Huber loss or SoftL1.

## 1 Introduction

Smartwatches, rings, and other wearable devices promise an era of preventive health and personalized fitness through continuous monitoring of physiological signals. Existing approaches to extract physiology from wearable data rely on least-squares fits and assume Gaussian noise [12]. However, real-world wearable data is often porous, noisy and irregular, including random drops, spikes, and other motion artifacts. These non-Gaussian disruptions prevent using traditional fitting methods to recover the underlying physiological parameters with sufficient precision to be useful. This limitation is currently overcome by focusing on the data recorded during sleep periods [13] where data is mostly free of noise and anomalies, or by averaging data for an entire day to reduce the impact of outliers [14]. Regressions that can cut through non-Gaussian noise are needed to analyze the entire day's data and extract additional valuable and actionable information from wearable data.

Traditional symbolic regression aims to infer both the structure and constants of symbolic equations. Here we address parameter recovery for known symbolic structures corrupted by non-Gaussian noise common in real-world wearable data, such as outliers, spikes and other artifacts.

For wearable data (e.g., heart rate, temperature, blood oxygen saturation), the symbolic structure of the physiological model is known (e.g., sinusoidal function for circadian rhythm), but the symbolic parameters need to be inferred from the noisy data. Recovering these constants directly, robustly, and accurately is essential for interpretability, personalization, and downstream modeling.

In this work, we introduce the Robust Regression Transformer (R2T), a novel hybrid neural-symbolic architecture for robust regression using a transformer encoder with a highly compressed output to process numerical physiological signal sequences and extract symbolic parameters which, when passed through a symbolic decoder, reconstruct the original input sequence. During training our model ingests synthetic time-series data which is intentionally corrupted with realistic wearable-device noise (e.g., patterned disruptions and spikes) and learns to reconstruct the underlying symbolic parameters for the known expression forms without supervision.

Unlike previous neural-symbolic regression methods [2][3][4], which try to recover full expressions using nearly perfect data, our method focuses on parameter estimation for known symbolic forms under structured noise and missing data. We show that training on synthetic clean vs. corrupted signal pairs enables R2T to robustly recover symbolic parameters with high precision.

Our experiments demonstrate that R2T consistently outperforms robust classical fits in recovering symbolic parameters under high corruption. This represents, to our knowledge, the first demonstration of transformer-based symbolic parameter recovery under non-Gaussian asymmetric noise, tailored to the challenges of wearable sensor data.

## 2 Background

### 2.1 Symbolic Regression and Parameter Estimation

General symbolic regression aims to infer both structure and constants of symbolic equations from data, using techniques like genetic programming [11] and neural methods:

- Deep Learning for Symbolic Mathematics [2] used transformers for seq2seq machine translation to solve integration and differential equations from symbolic representations using a new syntax to represent mathematical expressions.
- End-to-end symbolic regression with transformers [3] introduced an end-to-end transformer that predicts full symbolic expressions, including constants, using BFGS to refine the constants.
- SymFormer [4] describes an improved approach for jointly predicting structure and numeric constants, using gradient-based SGD fine-tuning for constant refinement.

These works train using clean or lightly perturbed symbolic function data, but they don't address parameter recovery for known symbolic structures corrupted by non-Gaussian noise common in wearable data. They focus on recovering the structure of the symbolic expression, using standard methods to recover the constants. However, BFGS, SGD (or any other approach that recovers symbolic constants by minimizing the least squares error between the predicted formula's outputs and the observed signals) will produce the wrong fit when the noise is structured, temporally variant and not symmetric.

### 2.2 Robust Regression

Classical robust regression techniques, such as Huber loss [5] and SoftL1 [6] are designed to reduce the influence of outliers but are limited when facing dense, structured, asymmetric noise, including clustered outliers and repeated spikes. These robust regression techniques still use least squares and fail when the data has structured asymmetric noise. Representation Learning for Wearable-Based Applications in the Case of Missing Data [7] used a transformer for imputing missing wearable data better than other interpolation techniques. However, this work did not address the issue of removing spikes and bad data and does not provide a path to symbolic parameter extraction. Neural denoising models trained on synthetic noise patterns are a promising alternative but have not been extended to non-Gaussian noise or adapted to symbolic parameter extraction [8].

### 2.3 Self-Supervised Learning via Masking

Enhanced training via masked modeling has been shown to impart strong generalization capabilities to transformers (e.g., BERT [9]) and has been successfully adapted for symbolic tasks (e.g., SymFormer masks constants during training [4]). To our knowledge, our work is the first to leverage masked pretraining on corrupted wearable physiological signals to recover symbolic parameters from known expression forms.

# 3 Model Architecture

The proposed neural network architecture (Figure 1) is based on the original multi-layer transformer encoder architecture by Vaswani et al. [1], but we don't use the decoder portion of the architecture. An encoder alone can handle tasks such as classification, regression, and even generative tasks such as filling gaps in the data, so the decoder is not really needed, while using about half the model parameters when compared with the complete transformer. The transformer encoder generates a vector that encodes the meaning and context of the input data sequence by performing the following operations: (i) The input sequence is converted into a sequence of embeddings, (ii) they are added to position embeddings to generate a sequence of input vectors, (iii) the multi-headed self-attention block takes in these vectors as query (Q), keys (K) and values (V) for each layer of the Encoder to perform data sequence context interpretation using vector dot products, (iv) the new vectors are added to the residual input vectors and normalized before entering a multilayer neural network (feed forward network) to perform additional operations, and (v) the output of the feed forward network is added to its input residual and normalized again.

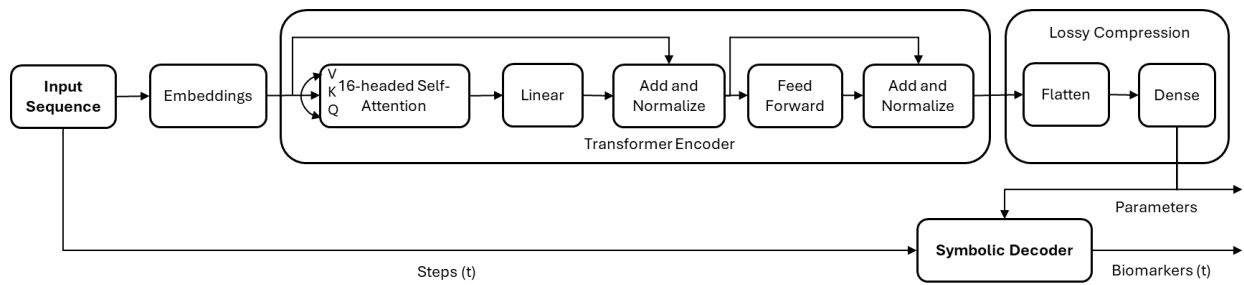

*Figure 1. The R2T model architecture.*

## 3.1 Lossy Compression

The transformer encoder output is flattened and compressed into a one-dimensional tensor with only 9 constants by the Lossy Compression NN, which has a single dense layer with linear activation. Corresponding to the 9 symbolic parameters that are extracted from the data, these constants have physiological interpretations and can be used as biomarkers for health/wellness classification or other tasks.

## 3.2 Steps processing

Wearable devices track the number of steps taken as an indication of physical activity. The transformer uses the steps sequence along with the other physiological signal sequences to determine the parameters in the symbolic regression, so steps are naturally included in the transformer's input sequence. Since steps are minimum while sleeping and transformers lack an inductive bias for temporal smoothing, a smoothed version of steps is added to the input sequence to help the transformer determine the phase of the Circadian rhythm.

The symbolic decoder also needs the steps sequence to reconcile the changes in physiological signals when walking or running. For example, when a person is running, their heart rate is higher than when they are walking or sitting down. To focus the transformer encoder on the task of predicting symbolic parameters and avoid having to recreate the steps sequence, the input steps sequence is fed directly into the symbolic decoder.

## 3.3 Symbolic decoder

The symbolic parameters are used to reconstruct the physiological signal sequence using a Symbolic decoder, which uses equations that describe the physiological human body response to steps and change over time. These equations can be non-linear and with multiple fit parameters. In our case, they were chosen to closely fit the experimental data collected from different wearables. For example, for heart rate sequences, we used:

$$HR(t) = RHR + A_{HR} \sin\left(\frac{2\pi t}{T} + \varphi\right) + B_{HR} \cdot \tanh s(t) + C_{HR} \cdot \tanh s(t-1)$$

Here, $HR(t)$ is the heart rate sequence, $RHR$ is the resting heart rate, $A_{HR}$ is the Circadian rhythm amplitude, $t$ is the time sequence $\{0, 1, \ldots, 95\}$, $T$ is the sequence length (96), $\varphi$ is the phase of the Circadian rhythm, $B_{HR}$ is the sensitivity of heart rate to steps, $s(t)$ is the scaled steps sequence. $C_{HR}$ is the delayed steps sensitivity, and $s(t-1)$ is the scaled steps delayed by one time increment (15 min). While not previously used in wearables, the steps sensitivity and delayed steps sensitivity are useful parameters to determine a person's physical fitness. For example, when the body is fighting an infection, there can be a change in how the heart rate responds to exercise and how quickly it recovers, even before there are other symptoms.

### 3.4 Synthetic data generation

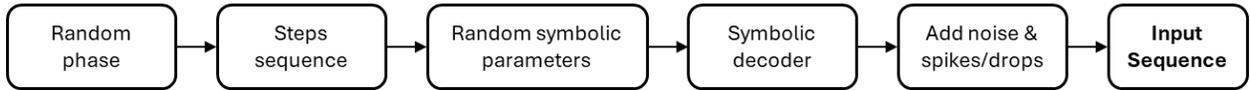

*Figure 2. Synthetic data generation flow.*

The symbolic decoder is also used to generate synthetic data for training, validation and testing using the following synthetic data generation flow (Figure 2):

- The circadian rhythm is the body's natural, internal process that regulates the sleep-wake cycle and repeats roughly every 24 hours. The circadian phase is randomly selected between 0 and $2\pi$ using an even distribution to ensure the transformer learns to fit the wearable data regardless of sleeping schedule.
- The steps sequence locations and amplitudes are then randomly selected, ensuring there are no steps while sleeping.
- The rest of the symbolic parameters are randomly selected using an even distribution with minimum and maximum values extracted from wearable data.
- The symbolic decoder uses these parameters and the steps sequence to generate the synthetic physiological signal sequences for temperature, heart rate and blood oxygen saturation.
- To generate synthetic data that closely mimics real-world data, we randomly remove chunks of data and add random noise, spikes and drops to simulate signal fluctuations caused by sensor noise, motion artifacts, stress, wearable removal, and other causes.
- These realistic synthetic sequences are used as input sequences.

Generating synthetic data in this manner is advantageous for a few reasons. First, it provides noiseless targets to guide the training, enabling symbolic regressions of signals with any type of noise regardless of structure or symmetry. Second, many sequences can be easily generated and used for training to achieve good generalization. Finally, once trained and used on real wearable data, the statistical distribution of symbolic parameters provided by R2T can be used to generate new training data and recursively improve performance on real-world data.

### 3.4 Data processing, normalization and masking

Since the signals collected by wearable devices have different sample rates, we elected to average the data using 15-minute intervals. If all data is missing during a 15-minute interval, the value is set to 0.0. The data is averaged and normalized between 0.5 and 1.0 corresponding to the minimum and maximum values for each physiological signal. For temperature, the minimum is 28 C and the maximum is 45 C (17 C range). Heart rate ranges from 30 beats per minute (bpm) to 220 bpm (190 bpm range). For SpO2 or blood oxygen saturation, 70% saturation is the minimum and 100% saturation is the maximum (30% saturation range). After normalization, the data sequence is masked, randomly setting 10% of values to 0.0, the same value used for missing data. The valid data values are numerically separated from masked/missing value by at least 0.5 to help differentiate missing data from regular data in the embedded vector space.

### 3.5 Model size

The size of the model can be quite small and still achieve excellent performance. For example, our initial model uses a single transformer layer with 16 attention heads, an embedding dimension of 192, and a compression output of 9 parameters for a total of 417,133 trainable parameters (1.59 MB).

## 4 Training

Training with synthetic data is unsupervised and no labeling is required. The signals coming out of the symbolic encoder used in data synthesis are free of noise or any other artifacts, so they are used as training targets. The input sequence with structured noise is masked, embedded using an MLP, added to sinusoidal positional encoding, and fed into the transformer encoder. The output symbolic parameters are compared with the original synthetic symbolic parameters to calculate a symbolic parameter mean-squared-error (MSE). The output of the symbolic decoder is compared with the target sequence to calculate a sequence reconstruction MSE. The combination of these MSEs is used as the loss to adjust the transformer encoder and lossy compression weights using gradient descent (Adam optimizer).

We used a masking-percentage of 0.1-0.3 and a masked value of 0.0 before embedding, which is the same value used for missing data. This encourages the model to learn to fill in missing data. We trained using 32,768 distinct randomly masked sequences. Each sequence contained five parameters corresponding to the three chosen physiological signals (temperature, heart rate, and oxygen saturation), steps and a smoothed version of the steps. We trained for 1,000 epochs with a custom callback to stop at the lowest validation MSE with a patience of 100 epochs, which triggered just before completing 700 epochs. Initial training rate is 0.001 and gradually reduced the learning rate using cosine schedule [10] to 0.0001 over 1,000 epochs. For regularization, we used dropout rate of 0.1 and clipnorm of 1.0.

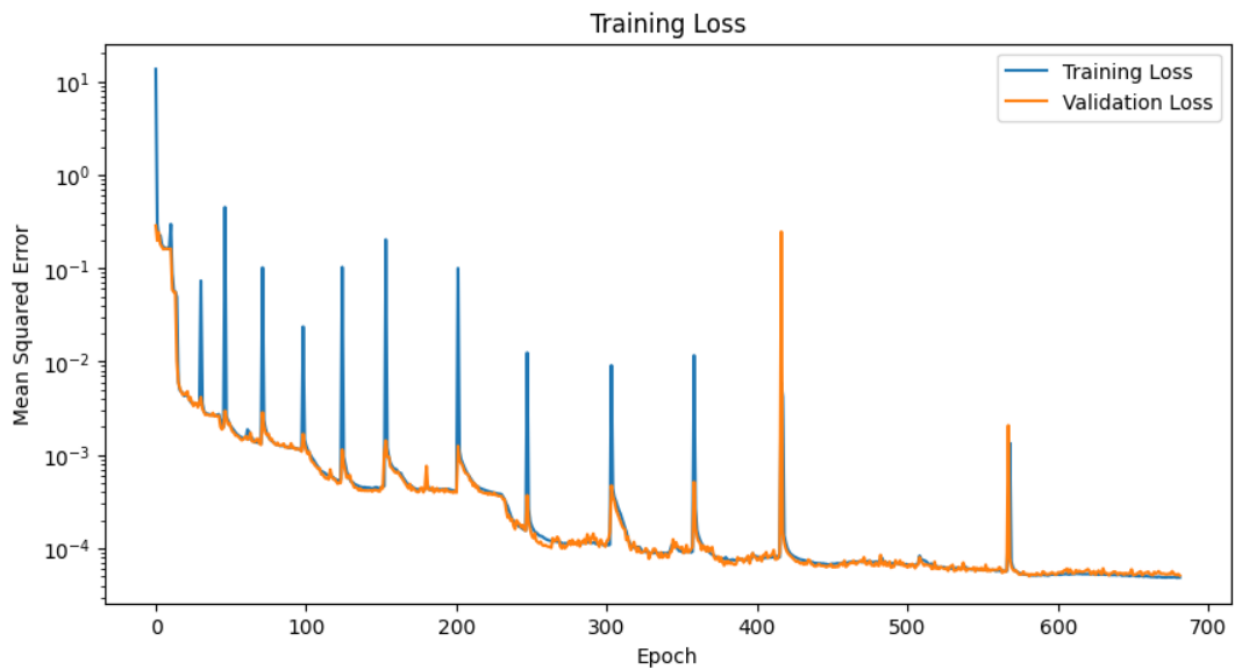

*Figure 3. Training curve with training and validation loss for each epoch*

We observe transient spikes in training and validation loss throughout training (Fig. 3). These may be caused by dropout-induced variance during forward passes, temporary gradient suppression from norm-based clipping, and the nonlinear operations in the symbolic decoder, which can exacerbate small deviations in the transformer outputs. Despite these fluctuations, the overall training curve exhibits stable downward trends for both training and validation losses, and the model achieves excellent final performance and generalization.

### 4.1 Loss calculation

The loss used during training is a weighted MSE computed from the sequence reconstruction MSEs for temperature ($MSE_T$), heart rate ($MSE_{HR}$), and blood oxygen saturation ($MSE_{SpO2}$) added to the symbolic parameter MSEs for circadian phase ($MSE_{CP}$), maximum temperature ($MSE_{MT}$) and resting heart rate ($MSE_{RHR}$) using:

$$training\ loss = 0.5 \cdot MSE_T + 2.0 \cdot MSE_{HR} + 0.5 \cdot MSE_{SpO2} + \frac{MSE_{CP}}{D_{CP}} + \frac{MSE_{MT}}{D_{MT}} + \frac{MSE_{RHR}}{D_{RHR}}$$

Here the D values dividing the symbolic parameter MSEs are dynamic denominators that change value between 1,000 and 10,000 to modify the importance of the symbolic parameter loss compared to the sequence reconstruction loss depending on the symbolic parameter MSE values:

$$D_i = 1000 \cdot 10^{\left(1.0 - \frac{min(1.0, max(0.1, MSE_i)) - 0.1}{0.9}\right)}$$

The average training loss and validation loss at the end of training was 5e-6, with a range of 9.14e-7 to 5.67e-4.

## 6 Results

The trained model was saved in Keras format and used for regression using both synthetic data and real wearable data. Here we only discuss the results from synthetic data testing, as they provide quantifiable results on how well the fit matches the original noise-free signal. During inference, R2T is used to predict symbolic parameters for all physiological sequences. The symbolic decoder is used to compute the sequence reconstruction and compare it with the target sequence to compute a sequence reconstruction MSE for each of the physiological signals. The scipy library least_squares function was used running standard regressions on non-zero data (to avoid the effects of missing data), where 'linear' loss was used for OLS, 'huber' loss was used for Huber, and 'soft_l1' loss was used for SoftL1.

We evaluate the performance of the model by running inference using a synthetic testing dataset with 6,364 unmasked sequences. For evaluation, the mean square error (MSE) for each signal in the sequence (temperature, heart rate, and SpO2) is separately computed.

To determine the best and worst overall fits, we use the testing loss, which is a simplified version of the weighted MSE used for training:

$$testing\ loss = 0.5 \cdot MSE_T + 2.0 \cdot MSE_{HR} + 0.5 \cdot MSE_{SpO2}$$

To visualize the quality of the R2T regression, we graph input and reconstructed sequences together (Figure 3), along with ordinary least squares (OLS) fit for comparison.

The best-case reconstruction (Figure 3a) had a R2T fit error of the order of 1e-6 for all three signals. Since the temperature sequence has small Gaussian noise and no asymmetric noise, and the OLS fit code was written to reject any missing points (values equal to 0.0), the R2T and OLS MSEs are similar. The maximum temperature and temperature circadian amplitude are matched to within 0.2% of the total valid temperature range (0.002*17C = 0.034C) and the circadian phase to within 0.16% of the total phase range (0.0016*360 deg = 0.6 deg). However, as soon as the magnitude of the noise increases or anomalies are introduced, the OLS fit deviates significantly while R2T continues to generate valid regressions. For example:

- The HR reconstruction plot has an anomaly (a HR increase lasting about five time-steps) around the 30th time step of the sequence, but the fit is still good, even correctly fitting the spikes in HR corresponding to the steps sequence. The R2T predicted resting HR and circadian amplitude are matched to within 0.1% of the valid HR range (0.001*190bpm = 0.19 bpm). The R2T steps sensitivity is perfectly matched in this case, while the delayed steps sensitivity is matched to within 0.01. However, the OLS fit is pushed up by this anomaly, substantially increasing the resting HR and Circadian amplitude errors to about 4% (7.6 bpm).
- The SpO2 sequence has large asymmetric noise. The R2T predicted average SpO2 is matched to within 0.2% of the valid SpO2 range (0.002*30% = 0.06% saturation) and the circadian amplitude is matched to within 0.2% (0.06% saturation). Using OLS, the predicted average SpO2 and circadian amplitude are off by about 8.5% of the valid range (2.5% saturation).

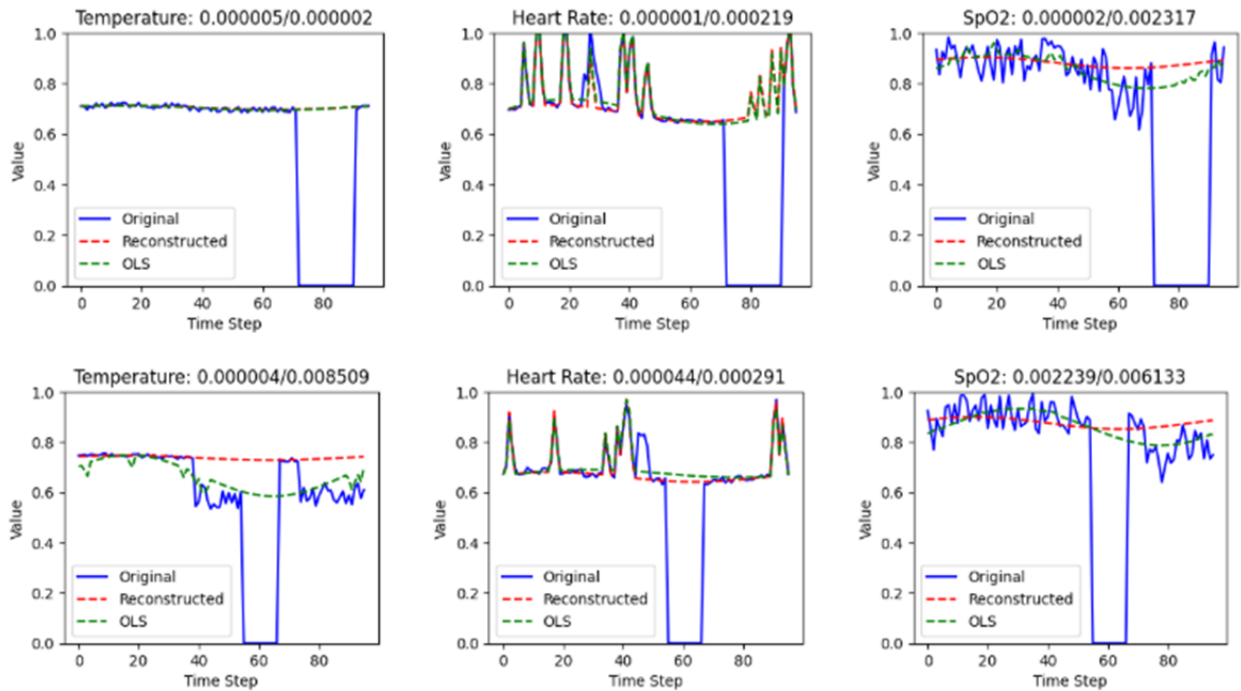

*Figure 3. (a-top) Best case fit and (b-bottom) worst-case fit using R2T. The plots show the original physiological sequences (temperature, heart rate, and SpO2) in blue, the R2T fit in red, and the OLS fit in green. Above each plot, the first number indicates the R2T loss, and the second number indicates the OLS loss..*

Most fits are close to the best-case, but there are some edge-cases. Here we look at the worst-case R2T reconstruction (Figure 3b), which had the worst overall testing loss with a R2T fit error of 4e-6 for temperature, 4e-5 for HR and 2e-3 for SpO2. For comparison, the OLS fit errors were 8.5e-3 for temperature, 2e-4 for HR and 6e-3 for SpO2. In this case, the R2T predicted maximum temperature and circadian amplitude are matched to within 0.2% (0.102 C) and circadian phase is matched to 3.5% (12.6 degrees). Note that phase is more challenging to fit due to its soft effect on the loss, particularly for small circadian amplitudes. The resting HR was matched to within 0.4% of full range (0.76 bpm) and the circadian amplitude to within 0.3% (0.57 bpm). Steps sensitivity was matched to within 0.03, and delayed steps sensitivity was matched to within 0.01. This was the largest testing loss from all sequences tested due to the SpO2 sequence fit error. The average SpO2 prediction was a bit low and matched to within 9.6% of full range (2.88% saturation) and the circadian amplitude matched to within 0.4% (0.12% saturation). In stark contrast, OLS failed to produce a good fit for any of the three sequences and the predicted parameters were off by 20% (3.4C) for temperature, 6% (11 bpm) for HR, and 20% (6% saturation) for SpO2.

In summary, the errors in R2T fit and symbolic parameter predictions are within acceptable measurement error for wearable devices, even for worst-case. However, OLS is only able to properly fit a sequence with small Gaussian noise, failing any time there is large noise or asymmetric anomalies, resulting in errors that are significantly above what is acceptable for wearable device measurements.

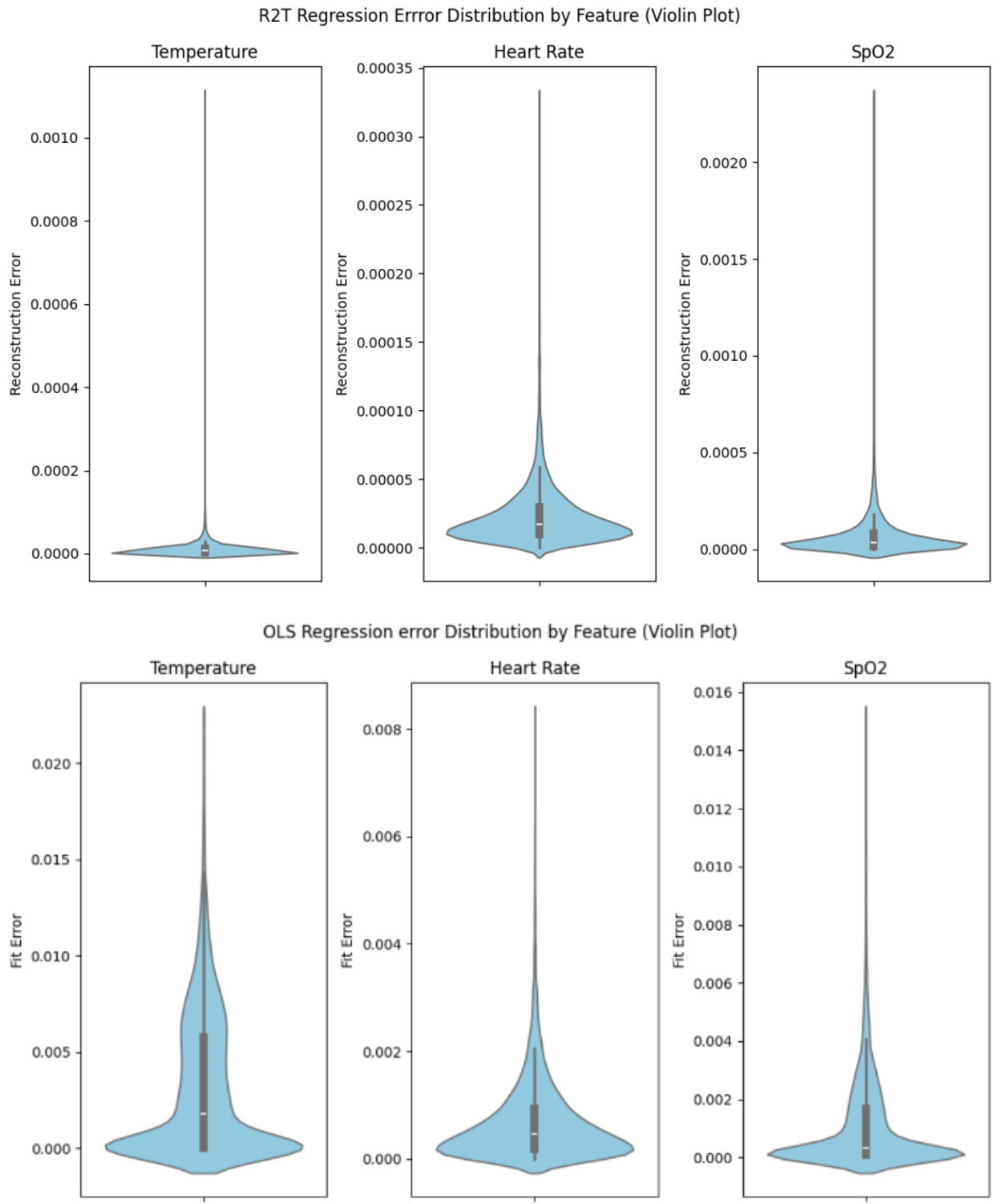

*Figure 4. Error distribution for each physiological signal using R2T (a-top) and OLS (b-bottom).*

Violin plots (Figure 4) help to visualize the statistical MSE distributions for all three physiological sequences to compare R2T and OLS regressions. The R2T regression has excellent performance and a tight distribution for all three physiological signals (Figure 4a) with a median MSE of 6e-6 for temperature, 1.7e-5 for heart rate and 3.5e-5 for SpO2. The OLS fit generates higher loss values and a much broader distribution for all three physiological signals (Figure 4b) with a median temperature MSE of 1.8e-3, median HR MSE of 4.6e-4 and median SpO2 MSE of 3.3e-4. All distributions have long tails, indicating the presence of a few edge cases that are difficult to fit, as previously described in Figure 3b. Note that due to auto-scaling, the y-axis range for the OLS violin plots is much larger than the y-axis range for the R2T violin plots.

Table 1 shows the MSE percentiles for each regression method and physiological signal, with the best performing (lowest MSE) in bold. R2T has the best fit performance in all cases for the data we evaluated. However, the improvement is not the same for all types of data. For temperature, which has small Gaussian noise with occasional drops and missing data, the median MSE improvement is about a factor of 300. For SpO2, which has a large Gaussian noise with occasional drops and missing data, the median MSE improvement is about a factor of 10. For HR, which has small Gaussian noise, spikes caused by steps, occasional increases of variable width (HR jumps caused by stress) and missing data, the median MSE improvement is about a factor of 27.

*Table 1. Mean square error for each physiological signal using our R2T regression compared with OLS, Huber and SolfL1 least-squares regressions.*

| Percentile | Temperature | | | | Heart Rate | | | | SpO2 | | | |
|---|---|---|---|---|---|---|---|---|---|---|---|---|
| | R2T | OLS | Huber | SoftL1 | R2T | OLS | Huber | SoftL1 | R2T | OLS | Huber | SoftL1 |
| 10% | **0.000001** | 0.000001 | 0.000001 | 0.000001 | **0.000005** | 0.000036 | 0.000036 | 0.000036 | **0.000006** | 0.000024 | 0.000024 | 0.000024 |
| 20% | **0.000002** | 0.000003 | 0.000003 | 0.000003 | **0.000008** | 0.000132 | 0.000132 | 0.000132 | **0.000011** | 0.000055 | 0.000055 | 0.000055 |
| 30% | **0.000003** | 0.000006 | 0.000006 | 0.000006 | **0.000011** | 0.000232 | 0.000232 | 0.000230 | **0.000017** | 0.000099 | 0.000099 | 0.000098 |
| 40% | **0.000005** | 0.000431 | 0.000431 | 0.000428 | **0.000014** | 0.000333 | 0.000333 | 0.000333 | **0.000025** | 0.000170 | 0.000170 | 0.000170 |
| 50% | **0.000006** | 0.001792 | 0.001792 | 0.001782 | **0.000017** | 0.000457 | 0.000457 | 0.000454 | **0.000035** | 0.000330 | 0.000330 | 0.000328 |
| 60% | **0.000008** | 0.003202 | 0.003202 | 0.003201 | **0.000021** | 0.000610 | 0.000610 | 0.000608 | **0.000048** | 0.000720 | 0.000720 | 0.000716 |
| 70% | **0.000011** | 0.004923 | 0.004923 | 0.004921 | **0.000026** | 0.000812 | 0.000812 | 0.000809 | **0.000067** | 0.001327 | 0.001327 | 0.001321 |
| 80% | **0.000015** | 0.006515 | 0.006515 | 0.006517 | **0.000034** | 0.001107 | 0.001107 | 0.001102 | **0.000100** | 0.002057 | 0.002057 | 0.002051 |
| 90% | **0.000024** | 0.008480 | 0.008480 | 0.008482 | **0.000047** | 0.001622 | 0.001622 | 0.001615 | **0.000168** | 0.003065 | 0.003065 | 0.003060 |
| 100% | **0.001105** | 0.021673 | 0.021673 | 0.021670 | **0.000326** | 0.008159 | 0.008159 | 0.008166 | **0.002331** | 0.014979 | 0.014979 | 0.014974 |

In conclusion, R2T outperforms all least-squares-based fits, especially when data contains large-amplitude Gaussian noise (e.g., compare 10[th] percentile for SpO2 sequence fit) or asymmetric noise (e.g., compare > 40[th] percentile for Temperature). When the noise is small and exclusively Gaussian (e.g., compare 10[th] percentile for Temperature sequence fit), R2T performs about the same as least squares. Finally, since the three least squares fit methods performed about the same, we conclude that non-linear loss for robust regression used by Huber and SofL1 do not improve performance significantly for real-world wearable data. On the other hand, the proposed R2T provides a reliable method for extracting useful information from such data.

# 7 Conclusion

In this work, we presented the R2T, a novel hybrid neural-symbolic model using transformers that shows dramatically improved regression performance for noisy and porous non-linear functions over traditional least-squares fit, including robust regression using non-linear loss.

The Robust Regression Transformer (R2T) can be used for myriad applications in data analysis, wherever least-squares fit is currently applied. Next steps are to improve synthetic data generation by recursively modifying the noise distribution to match the real-world data.

**Acknowledgments:** This work was supported by the Defense Health Agency (DHA) SBIR and the Defense Threat Reduction Agency (DTRA), Chemical & Biological Defense Program – CB11203.